\documentclass{article} % For LaTeX2e
\usepackage{baichuan}

\usepackage[utf8]{inputenc} % allow utf-8 input
\usepackage[T1]{fontenc}    % use 8-bit T1 fonts
\usepackage{hyperref}       % hyperlinks
\usepackage{url}            % simple URL typesetting
\usepackage{booktabs}       % professional-quality tables
\usepackage{amsfonts}       % blackboard math symbols
\usepackage{nicefrac}       % compact symbols for 1/2, etc.
\usepackage{microtype}      % microtypography
\usepackage{xcolor}         % colors
\usepackage{multirow}
\usepackage{multicol}
\usepackage{amsmath}
\usepackage{graphicx}
\usepackage{amsmath}

\usepackage{subcaption}
\usepackage{utfsym}
\usepackage{svg}
\usepackage{array}
\usepackage{graphicx}
\usepackage{fancyhdr}
\usepackage{tablefootnote}
\usepackage{footnote}
\usepackage{diagbox}
\usepackage{enumitem}
\usepackage{dsfont}
\usepackage{diagbox}

\title{\centering Full-ECE: A Metric For Token-level Calibration
on Large \\ Language Models}

\author{%
\quad \quad 	Han Liu$^{*}$ \\ 
	Tsinghua University\\
   \quad \space \space Baichuan Inc. \\
	\And
 Yupeng Zhang\thanks{Equal contribution} \\
	\hspace{0.5em} Baichuan Inc. \\
	\And
Bingning Wang$^{\dagger}$  \quad\quad\\
	\quad  Baichuan Inc. \\ 
	\AND
	\quad\quad\quad\quad\quad\quad\quad\quad Weipeng Chen \\
	\quad\quad\quad\quad\quad\quad\quad\quad \space \space Baichuan Inc. \\
	\And
	\quad \space Xiaolin Hu\thanks{Corresponding author, \texttt{daniel@baichuan-inc.com, xlhu@tsinghua.edu.cn}}\quad\quad\quad\quad\quad\\
	Tsinghua University \quad\quad\quad\quad\quad\quad\quad\\
	\And
	\quad\quad\quad\quad\quad\quad\quad\quad\quad\quad\quad\quad\quad\quad\quad   \\
	\quad\quad\quad\quad\quad\quad\quad\quad\quad\quad\quad\quad\quad\quad\quad \space  
}

\colmfinalcopy

\begin{document}

	\maketitle

	\begin{abstract}\label{abstract}
		Deep Neural Networks (DNNs) excel in various domains but face challenges in providing accurate uncertainty estimates, which are crucial for high-stakes applications. Large Language Models (LLMs) have recently emerged as powerful tools, demonstrating exceptional performance in language tasks. However, traditional calibration metrics such as Expected Calibration Error (ECE) and classwise-ECE (cw-ECE) are inadequate for LLMs due to their vast vocabularies, data complexity, and distributional focus. To address this, we propose a novel calibration concept called \textit{full calibration} and introduce its corresponding metric, \textit{Full-ECE}. Full-ECE evaluates the entire predicted probability distribution, offering a more accurate and robust measure of calibration for LLMs.
	\end{abstract}
	\section{Introduction}\label{introduction}
	
	Deep Neural Networks(DNNs) have achieved remarkable success across various domains, demonstrating superior performance in many tasks. Despite these impressive achievements, a critical challenge remains: the calibration of these models' predictions. Calibration refers to the model’s ability to provide uncertainty estimates that accurately reflect the true likelihood of its predictions being correct. In many high-stakes applications such as healthcare~\cite{leibig2017leveraging,dolezal2022uncertainty}, self-driving ~\cite{michelmore2020uncertainty}, and protein engineering~\cite{greenman2023benchmarking}, it is not sufficient for a model to be highly accurate; it must also provide reliable estimates of uncertainty to ensure safety and robustness.

Large Language Models~\cite{achiam2023gpt,touvron2023llama,brown2020language}  have recently emerged as powerful tools in the DNN area, demonstrating exceptional performance in a wide array of language tasks. It is essential to ensure that these models are well-calibrated at the token level. Token-level calibration refers to the alignment of the predicted probability distribution for each token with the true distribution observed in the corpus. This ensures that the probabilities assigned to each token reflect their actual occurrence likelihoods.
Token-level calibration differs from traditional classification task calibration in several aspects:
\begin{enumerate}[itemsep=0.2pt]
\item \textbf{Vocabulary Size}: Token-level calibration deals with a vast number of classes, typically in the hundreds of thousands according to the vocabulary size, vastly surpassing traditional classification tasks.

\item \textbf{Data Complexity and Imbalance}: LLMs are trained on datasets containing tens of billions of tokens, introducing significant complexity and diversity compared to standard datasets. Moreover, token-level calibration grapples with highly imbalanced token distributions, posing additional challenges.

\item \textbf{Distribution Focus}: Traditional calibration methods focus on the top-1 prediction. In contrast, token-level calibration for LLMs must account for the entire probability distribution across all tokens, as inference involves sampling from this distribution, making every token's probability significant.
\end{enumerate}

These distinctions underscore the inadequacy of traditional calibration metrics like Expected Calibration Error (ECE)~\cite{naeini2015obtaining} and classwise-ECE (cw-ECE)~\cite{kull2019beyond} when applied to LLMs. ECE primarily considers the probability of the most likely token, which, although suitable for traditional classification tasks, fails to capture the nuances of the token distribution in LLMs. In contrast, cw-ECE evaluates calibration across each class but struggles with the extreme class imbalance and the sheer number of classes in LLMs. As the vocabulary size increases, the frequency of many tokens diminishes, making it challenging to obtain reliable calibration metrics based on these rare tokens. 
%Consequently, cw-ECE values tend to diminish significantly, rendering this metric less useful for LLMs.

To address these challenges, we propose a novel calibration concept called \textit{full calibration} and introduce its corresponding metric, \textit{Full-ECE}. Full-ECE evaluates the calibration of the entire predicted probability distribution, encompassing all tokens rather than focusing on the top-1 prediction. This comprehensive approach ensures that the calibration metric is robust to imbalances and vast class count of LLMs, providing a more accurate measure of LLMs' calibration. 

\begin{figure*}[t]
  \centering
  \includegraphics[width=0.9\textwidth]{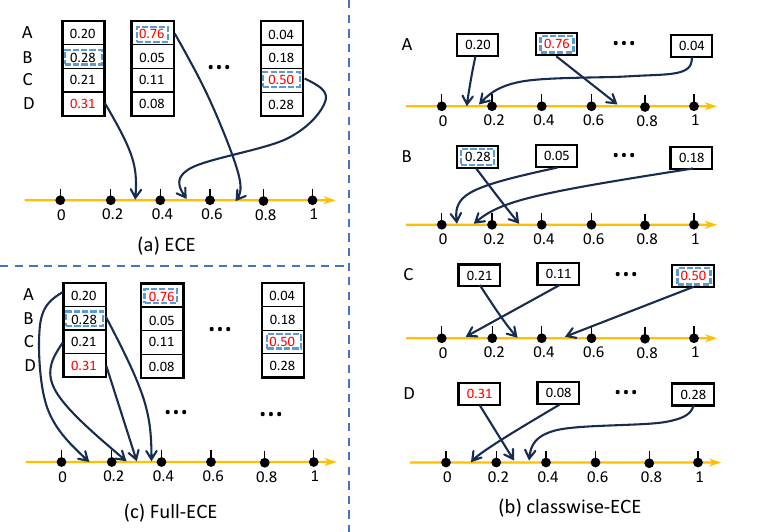}
  \caption{\small{(a), (b), and (c) demonstrate how ECE, cw-ECE, and Full-ECE aggregate statistics per bin for a 4-class classification task. Red numbers represent maximum probability values, and blue dashed boxes indicate ground truth label categories. ECE considers only these red maximum values, cw-ECE computes ECE for each class individually, while Full-ECE enhances cw-ECE by combining bins across different classes for statistical analysis.}}
\label{fig:eces}
\end{figure*}
	
	\section{Background}
	Calibration refers to the process of adjusting the predictive probabilities of a model so that they reflect the true likelihood of outcomes. For example, if a model predicts an event with a probability of 70\%, that event should occur approximately 70\% of the time in the long run. 
The existing concepts of calibration can mainly be divided into two types: confidence calibration and classwise calibration. 

To introduce these two types of calibration, we first need to explain some fundamental concepts.
Let $P_{data}$ denote the probability distribution of data. A dataset consists of finite $(x,y)$ pairs where $x\sim P_{data}$ is a data instance and $y \in \mathcal{Y}=\{1,2,...,K\}$ is the ground-truth class label of $x$.
A model takes in the instance $x$ and output its prediction vector $p\in \mathbb{R}^{K}$. Let $p^{(k)}$ denote the $k$-th dimension of $p$.

\subsection{Confidence Calibration}
In confidence calibration, the index corresponding to the maximum value in the model's predicted probability distribution $\hat{y}=\text{argmax}_{k\in \mathcal{Y}}\{p^{(k)}\}$ is taken as the model's prediction, and the associated probability value $\hat{p}=\text{max}_{k\in \mathcal{Y}}\{p^{(k)}\}$ is considered as the model's confidence. A model is confidence-calibrated if for any confidence $c \in [0,1]$, 
\begin{equation}
\begin{aligned}
\mathbb{P}(\hat{y}=y|\hat{p}=c)=c.
\label{equ:conf calibration}
\end{aligned}
\end{equation}
A widely used metric for evaluating confidence calibration is the ECE~\cite{naeini2015obtaining}, which is defined as the expected absolute difference between the model’s confidence and its accuracy: 
\begin{equation}
\begin{aligned}
{\text{ECE}}=\mathbb{E}_{\hat{p}}|\mathbb{P}(\hat{y}=y|\hat{p})-\hat{p}|.
\label{equ:ece}
\end{aligned}
\end{equation}
Since we only have finite samples, ECE cannot be directly calculated using the definition provided above. In practical calculations, we replace the above definition with a discretized version of ECE in which the interval $[0,1]$ is divided into $M$ equispaced bins. Let $B_m$ denote the indices of samples with confidences belonging to the $m$-th bin (i.e. $(\frac{m-1}{M},\frac{m}{M}]$). The accuracy of this bin is $A_m=\frac{1}{|B_m|}\sum_{i\in B_m}\mathds{1}(\hat{y}_i=y_i)$. The average confidence of this bin is $C_m=\frac{1}{|B_m|}\sum_{i\in B_m}\hat{p}_i$. The discretized version of ECE is defined as
\begin{equation}
\begin{aligned}
{\text{ECE}}=\sum_{m=1}^M\frac{|B_m|}{N}|A_m-C_m|,
\label{equ:ECE_finite}
\end{aligned}
\end{equation}
where $N$ is the number of samples in the dataset.
\subsection{Classwise Calibration}
Classwise calibration is concerned with calibrating the predicted probabilities for each individual class in a multi-class classification problem. It ensures that the predicted probabilities across different classes are accurate and properly calibrated. A model is classwise-calibrated if for any class $k$ and any probability $q$,
\begin{equation}
\begin{aligned}
\mathbb{P}(y=k|p^{(k)}=q)=q.
\label{equ:classwise calibration}
\end{aligned}
\end{equation}
Similar to ECE for confidence calibration, the cw-ECE~\cite{kull2019beyond} for classwise calibration is defined as
\begin{equation}
\begin{aligned}
{\text{cw-ECE}}=\frac{1}{K}\sum_{k=1}^K {\text{class-}}k{\text{-ECE}},
\label{equ:cw-ece}
\end{aligned}
\end{equation}
where ${\text{class-}}k{\text{-ECE}}$ is computed on the $k$-th class:
\begin{equation}
\begin{aligned}
{\text{class-}}k{\text{-ECE}}=\mathbb{E}_{p^{(k)}}|\mathbb{P}(y=k|p^{(k)})-p^{(k)}|.
\label{equ:class-k-ece}
\end{aligned}
\end{equation}

The discretized version of cw-ECE is defined as
\begin{equation}
\begin{aligned}
{\text{cw-ECE}}=\sum_{k=1}^K\sum_{m=1}^M\frac{|B_{m,k}|}{NK}|A_{m,k}-C_{m,k}|,
\label{equ:classwise ECE_finite}
\end{aligned}
\end{equation}
where $B_{m,k}$ denotes the set of indices of samples whose predicted probabilities of the $k$-th class lie in the $m$-th bin,  $A_{m,k}=\frac{1}{|B_{m,k}|}\sum_{i\in B_{m,k}}\mathds{1}(y_i=k)$ and $C_{m,k}=\frac{1}{|B_{m,k}|}\sum_{i\in B_{m,k}}p^{(k)}_i$.
	
\section{Full Calibration: Calibrating the Whole Distribution Predicted by LLMs}
As mentioned earlier, ECE considers only the calibration of the most probable class in the probability distribution, while classwise-ECE (cw-ECE) takes into account the calibration of all classes in the probability distribution. However, as shown in Equations \eqref{equ:cw-ece}
and \eqref{equ:classwise ECE_finite}, for each class $k$, cw-ECE requires calculating its corresponding ECE metric. ECE is a statistical metric that requires a sufficient number of samples for each class to be calculated accurately. In token-level calibration, it is impossible to ensure that each token appears frequently enough in the test set. 
We analyzed the distribution of token occurrences in a test set containing 5000 sentences and found that 30\% of the tokens do not appear in the dataset at all, and 41\% of the tokens appear only 1-10 times. This indicates that 71\% of the tokens appear fewer than 10 times, making cw-ECE evaluation inaccurate.

To address the issues of existing calibration metrics, ECE and cw-ECE, in token-level calibration, we propose a new and more suitable calibration concept: full calibration, along with its corresponding calibration metric, Full-ECE.

Unlike confidence calibration, where the index corresponding to the maximum value in the model's predicted probability distribution is taken as the model's prediction, full calibration views the model's output as a process of sampling from the predicted probability distribution. The sampled index $y^*$ is taken as the output token, and its corresponding probability value $p^*$ is considered as its confidence. A model is full-calibrated if for any $q \in [0,1]$, the following holds:
\begin{equation}
\begin{aligned}
\mathbb{P}(y^*=y|p^*=q)=q.
\label{equ:full calibration}
\end{aligned}
\end{equation}
Naturally, similar to the definition of ECE, we can define the metric corresponding to full calibration as Full-ECE:
\begin{equation}
\begin{aligned}
{\text{Full-ECE}}=\mathbb{E}_{p^*}|\mathbb{P}(y^*=y|p^*)-p^*|.
\label{equ:full-ece}
\end{aligned}
\end{equation}
According to Bayes' theorem and the law of total probability,
\begin{equation}
\begin{aligned}
\mathbb{P}(y^*=y|p^*)&=\frac{\mathbb{P}(y^*=y,p^*)}{\mathbb{P}(p^*)} \\
&=\frac{\sum_{k=1}^K \mathbb{P}(y^*=y=k,p^*)}{\sum_{k=1}^K \mathbb{P}(y^*=k,p^*)}.
\label{equ:full-ece2}
\end{aligned}
\end{equation}
Similar to the previously defined discrete ECE and cw-ECE, we divide the probability interval [0,1] of each class $k$ into $M$ equally spaced bins.
Each bin $B_{m,k}$ represents the set of indices of samples whose predicted probabilities of the $k$-th class lie in the $m$-th bin. 
Let $|B^*_m|$ represent the sum of the quantities in the 
$m$-th bin $B_{m,k}$ for all classes $k$, i.e. $|B^*_m|=\sum_{k=1}^K |B_{m,k}|$.
Base on~\eqref{equ:full-ece} and~\eqref{equ:full-ece2}, Full-ECE can be discretized and represented as
\begin{equation}
\begin{aligned}
{\text{Full-ECE}}&=\sum_{m=1}^M \frac{|B^*_m|}{N}|A^*_m-C^*_m|,
\label{equ:full-ece_finite}
\end{aligned}
\end{equation}
where 
\begin{equation}
\begin{aligned}
A^*_m=\frac{\sum_{k=1}^K \sum_{i \in B_{m,k}} \mathds{1}(y_i=k)}{|B^*_m|},
\label{equ:full-ece_A}
\end{aligned}
\end{equation}
and
\begin{equation}
\begin{aligned}
C^*_m=\frac{\sum_{k=1}^K \sum_{i \in B_{m,k}} p_i^{(k)}}{|B^*_m|}.
\label{equ:full-ece_C}
\end{aligned}
\end{equation}
Figures~\ref{fig:eces} depicts the aggregation methods of ECE, cw-ECE, and Full-ECE for each bin. ECE focuses solely on the highest probability value within each distribution, whereas cw-ECE considers the distribution across all classes $k$, while Full-ECE combines the statistics of different classes within the same bin. This approach addresses the issue in token-level calibration where cw-ECE faces the challenge of having too few samples for many classes.

\section{The Robustness of Full-ECE}
\begin{table}[htbp]
  \centering
  \begin{tabular}{lcc}
    \hline
     & \textbf{2B params} & \textbf{7B params}\\
    \hline
    RSD of cw-ECE     &   40.93\%  &   44.90\%   \\
    RSD of Full-ECE      &  7.56\%    &  8.36\%    \\
    \hline
  \end{tabular}
  \caption{The RSD of Full-ECE and cw-ECE for different models as $M$ varies.}
  \label{tab:rsd}
\end{table}
Both Full-ECE and cw-ECE are metrics designed to evaluate the calibration of the entire predicted distribution. 
From their definitions in the discrete case, it can be observed that both of them require the probability interval [0,1] to be divided into $M$ equal-length bins. A robust metric should exhibit lower relative variation as the value of $M$ changes. We evaluated the relative standard deviation (RSD) of Full-ECE and cw-ECE for different values of $M$ in the set $\{5, 10, 20, 50, 100, 200, 500\}$. Our experiments were conducted on two models: a 2-billion parameter GPT model and a 7-billion parameter GPT model trained on 1 trillion tokens. The lower the RSD, the more stable the metric is.
The experimental results, shown in Table~\ref{tab:rsd}, demonstrate that for both models, the RSD of Full-ECE as $M$ varies is significantly lower than that of cw-ECE. Therefore, Full-ECE is more stable across different values of $M$ compared to cw-ECE, highlighting its advantage in evaluating token-level calibration in large language models.
\section{Continuous Improvement of Full-ECE during LLM Training}
\begin{figure}[htbp]
  \centering
  \includegraphics[width=0.5\textwidth]{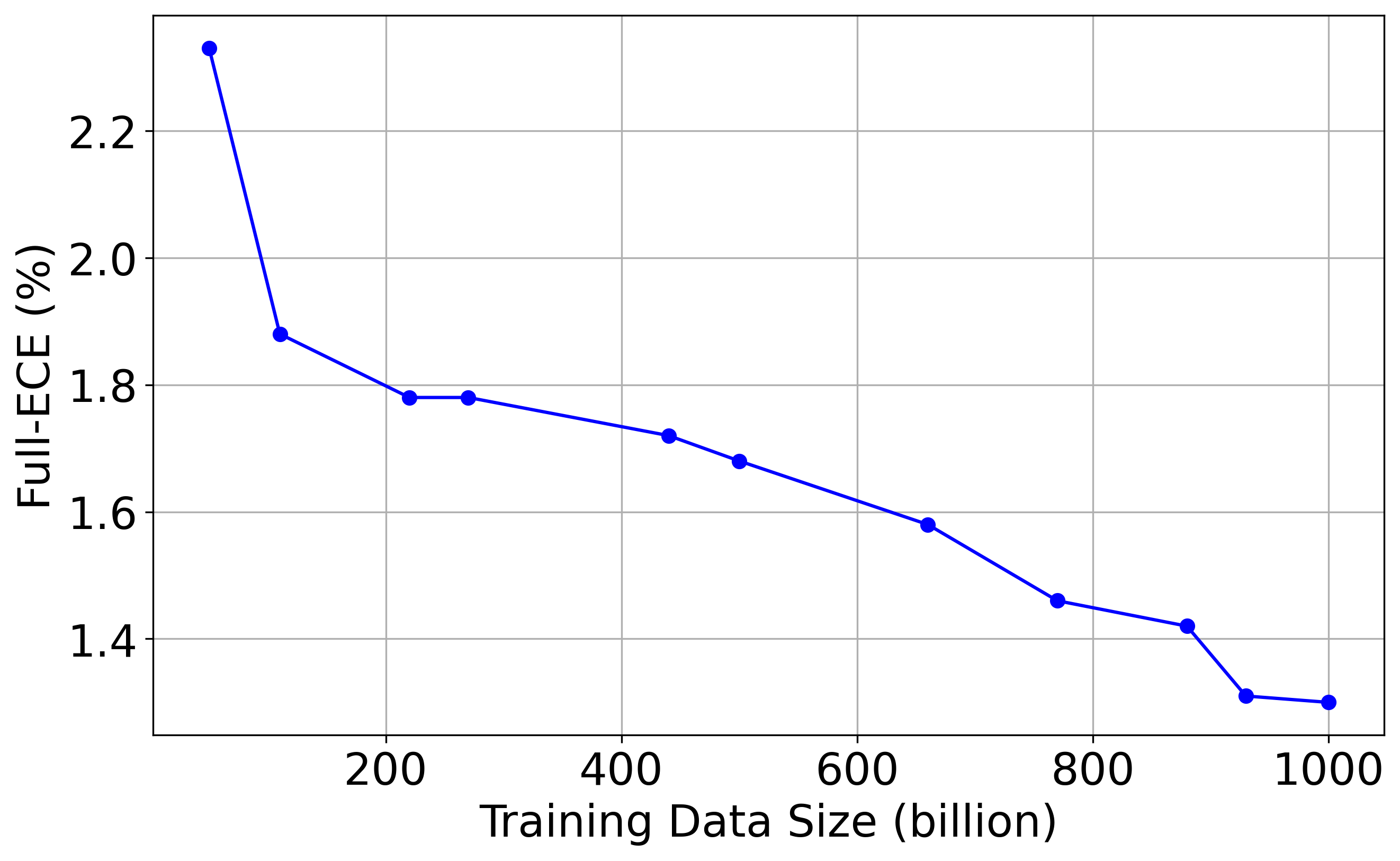}
  \caption{\small{Full-ECE with varying training data}}
\label{fig:training data}
\end{figure}
Previously, we demonstrated across different dimensions that Full-ECE is a more suitable metric than ECE and cw-ECE for evaluating uncertainty in LLMs. Another criterion for effective model evaluation is reliability and discriminability, where metrics should consistently improve with model capability. We tested the Full-ECE metric ($M=10$) at different training stages of the Baichuan-2 7B\cite{yang2023baichuan} model and observed a consistent downward trend (shown in Figure~\ref{fig:training data}), indicating a continuous improvement in token-level calibration throughout training.
\section{Conclusion}
We introduce the concept of full calibration and propose its corresponding metric, Full-ECE, to address the challenges in calibrating LLMs. Unlike traditional metrics such as ECE and cw-ECE, Full-ECE evaluates the entire predicted probability distribution across all tokens, From both experimental and theoretical perspectives, we have demonstrated that Full-ECE provides a more robust calibration measure for LLMs operating with vast vocabularies and complex data distributions.

		%%%%%%%%%%%%%%%%%%%%%%%%%%%%%%%%%%%%%%%%%%%%%%%%%%%%%%%%%%%%
		%		\clearpage
		%		\newpage
		\bibliography{baichuan}
		\bibliographystyle{baichuan}
		
		\newpage
		\appendix

	\end{document}